# Red AI? Inconsistent Responses from GPT Models on Political Issues in the US and China

Di Zhou[1], Yinxian Zhang[2]*

**Abstract**

The rising popularity of ChatGPT and other AI-powered large language models (LLMs) has led to increasing studies highlighting their susceptibility to mistakes and biases. However, most of these studies focus on models trained on English texts. Taking an innovative approach, this study investigates political biases in GPT's multilingual models. We posed the same questions about high-profile political issues in the United States and China to GPT in both English and simplified Chinese, and our analysis of GPT responses revealed that the bilingual models' political "knowledge" (content) and the political "attitude" (sentiment) are significantly more inconsistent on political issues in China. The simplified Chinese GPT models not only tended to provide pro-China information but also presented the least negative sentiment towards China's problems, whereas the English GPT was significantly more negative towards China. This disparity may stem from Chinese state censorship and US-China geopolitical tensions, which influence the training corpora of GPT bilingual models. Moreover, both Chinese and English models tended to be less critical towards the issues of "their own," represented by the language used, than the issues of "the other." This suggests that GPT multilingual models could potentially develop a "political identity" and an associated sentiment bias based on their training language. We discussed the implications of our findings for information transmission and communication in an increasingly divided world.

**Significance Statement**

This research is one of the first studies that systematically investigate the cross-language political biases and inconsistencies in large language models (LLMs). We found that GPT models trained in different languages may develop "political identities" and the associated sentiment biases that make them more positive toward their "own country" while more negative toward "other countries." In addition, we found that China-related political issues have significantly higher rates of inconsistency both in terms of content and sentiment, suggesting that Chinese state censorship and US-China geopolitical tensions may have influenced the performance of the bilingual GPT models. Our study brings public attention to the biases and inconsistencies in multilingual LLMs, which have profound implications for cross-cultural communications.

---

[1] Department of Sociology, New York University, New York, NY 10013.
[2] Department of Sociology, CUNY Queens College, New York, NY 11367.
* All correspondence should be directed to yinxian.zhang@qc.cuny.edu

## Introduction

ChatGPT[3] has been described as a revolutionary tool that may fundamentally change how information is obtained and transmitted (Chow and Perrigo 2023; Thompson 2022). Instead of typing keywords in a search engine and staring at the tens of thousands of returned web pages, tools powered by large language models (LLMs) such as ChatGPT can search and summarize the information behind the scenes, and return only a paragraph of the exact information you ask for in natural human language – just like an attentive personal assistant in science fictions.

However, it has been found that the quality of the information provided by the AI assistant is prone to mistakes and biases. Studies show that LLMs including the GPT not only contain gender, ethnic, political, and cultural biases (Acerbi and Stubbersfield 2023; Abid, Farooqi and Zou 2021; Brown et al. 2020; Lucy and Bamman 2021; Hartmann et al. 2023; Santurkar et al. 2023) but may also be a super-spreader of misinformation or "hallucinated information" (Perez et al. 2022; Ji et al. 2023). These pioneering studies, however, have been exclusively focused on the GPT models trained on English corpora. Nonetheless, GPT is designed with multilingual capabilities, and the applications of GPT models have extended far beyond English-speaking populations. This leads to an under-researched yet important question about LLMs' biases – assuming different languages may represent distinct cultures with different or even contradictory views on the same social, political, or cultural issues, do multilingual GPT models provide largely coherent answers to the same questions, or do they transmit information and values that are inconsistent or even contradictory, depending on the language used? Moreover, if the answers are inconsistent, how is such inconsistency manifested, and what types of cross-cultural biases can be identified?

These questions have profound implications for human communications. GPT presents itself as an objective AI assistant that provides undistorted summaries based on a comprehensive search of all available information. This can make average users more credulous about the accuracy and quality of the information provided. Consequently, if GPT multilingual models exhibit a systematic inconsistency, especially on contentious social and political issues, users with different language/cultural backgrounds may be exposed to implicitly biased information and could be oriented toward different beliefs without realizing it. As GPT becomes an increasingly popular source of information and opinions, it may influence the way people learn about and understand the world, including politics at home and abroad. The cross-language inconsistencies in GPT models could reinforce existing conflicts and cultural gaps between different populations. This includes not only citizens from different countries but also people from different migrant communities in a multicultural country like the United States (US). Eventually, it could increase rather than reduce the barriers to communication and collaboration across different cultures, contributing to a more fragmented world.

To probe into the cross-language inconsistency in GPT models, this study chooses simplified Chinese[4] and English as the main languages of interest and compares GPT answers to the same political questions in the two languages. The reason to focus on these languages is two-fold. First, the two languages represent two vastly different political and cultural systems – the English training corpora of GPT came from English sources primarily generated by American internet users (Bianchi 2023; Brown et al. 2020; Dixon 2022; Similarweb 2022; Wikipedia 2023) and the Chinese texts largely came from Mainland China under the rule of the Chinese Communist Party (CCP).[5] Due to the growing geopolitical tensions between the two countries and their vastly different cultures, their people may hold contrasting views on the same political issues. To find out if such differences are reflected in LLMs, we focused on GPT bilingual models' responses to political issues in the US and China.

---

[3] ChatGPT stands for Chat Generative Pre-trained Transformer. It is a large language model–based chatbot developed by OpenAI. GPT is the language model behind the chatbot.

[4] Unless specified, the Chinese language models in this study all refer to simplified Chinese.

Second, GPT models were primarily trained on information scraped from web pages. However, due to the stringent political censorship in China, the Chinese training corpus may be politically biased. Therefore, comparing Chinese GPT answers to the English ones also allows us to gauge the impact of censorship in LLMs building on the "open internet". Together, by testing GPT models with political questions in English and Chinese, we can examine if geopolitical tensions and political censorship affect GPT's performance between languages.

This study asks, given the geopolitical tensions and China's state censorship, how do GPT models respond differently to political issues in the US and China when the input language is different. Based on the *Human Rights Reports* and other official documents published by the US and Chinese authorities in the past twenty years, we developed a list of 533 questions covering social and political issues in the two countries. We then asked GPT 3.5 models the same questions twice, in English and Chinese respectively, and we compared the bilingual answers in terms of their political "knowledge" (*answer content*) and political "attitude" (*answer sentiment*). In other words, we carried out an experiment with GPT where only the input language is manipulated and other factors are controlled for, then we compared the two groups of answers to estimate the level of inconsistency in GPT bilingual models. Finally, we discussed the sources and the implications of such inconsistency.

## Results

### *GPT bilingual models' knowledge about China is the least consistent among all.*

To systematically explore the variations in GPT answers across languages, we first developed an analytical framework along two dimensions. First, we differentiate between political questions and natural science questions. Political questions are those concerning social and political problems criticized in official documents published by the US and Chinese authorities targeting each other. Based on the topics covered in these authoritative commentaries, we developed 533 political questions, including 266 China-related questions and 267 US-related questions.

Natural science questions, on the other hand, are questions about natural phenomena, basic physics, and basic chemistry. Assuming that natural science questions typically have one and only one correct answer, which should be consistent across languages and cultures, they form a good benchmark to gauge the level of inconsistency in GPT models as a result of distinct values and knowledge embedded in different languages. In total, we compiled a list of 184 natural science questions and expected to see political questions associated with a higher level of inconsistency compared to natural science questions.

The second dimension differentiates fact-based questions from opinion-based questions. Fact-based questions ask about numbers, statistics, or past incidents with supposedly objective records,

---

[5] Simplified Chinese is primarily used in mainland China. While there are other countries and regions where Chinese is the official or dominant language, the Chinese population is much smaller, and/or they mainly use Traditional Chinese characters in writing.

whereas opinion-based questions ask about subjective evaluations or judgments on similar topics. For example, on the issue of inequality in China, the question "What is China's Gini coefficient?" is fact-based, whereas the question "How severe is income inequality in China?" is opinion-based. Among the 533 political questions, we balanced the proportions of fact- and opinion-based questions, such that both US- and China-related questions are half fact-based and half opinion-based. Natural science questions are all fact-based. Due to the nature of fact- versus opinion-based questions, we expected to see opinion-based questions associated with a higher level of inconsistency.

| Fact- or Opinion-based | Science (N=184) | US Political (N=267) | China Political (N=266) | Overall (N=717) |
|---|---|---|---|---|
| Fact | 184 (100%) | 136 (50.9%) | 134 (50.4%) | 454 (63.3%) |
| Opinion | 0 (0%) | 131 (49.1%) | 132 (49.6%) | 263 (36.7%) |

**Table 1 Distribution of Questions by Topic and Question Type (Fact/Opinion).** Natural science questions are fact-based only and their consistency rate is used as the benchmark.

To begin with, we examined how consistent GPT answers were in terms of their content. To what extent the facts and opinions in GPT answers are consistent between the two languages? Note that content consistency differs from sentiment consistency. Two answers may present similar content but with contrasting undertones which, in turn, reveal different *attitudes*.

The two authors hand-coded the consistency of GPT bilingual answers to the 184 science questions and 533 political questions (question N = 717, answer N = 1,434, see coding details in Data and Methods). We reached a substantial inter-coder agreement rate with a Cohen's Kappa of 0.80. Figure 1 below summarizes the results.

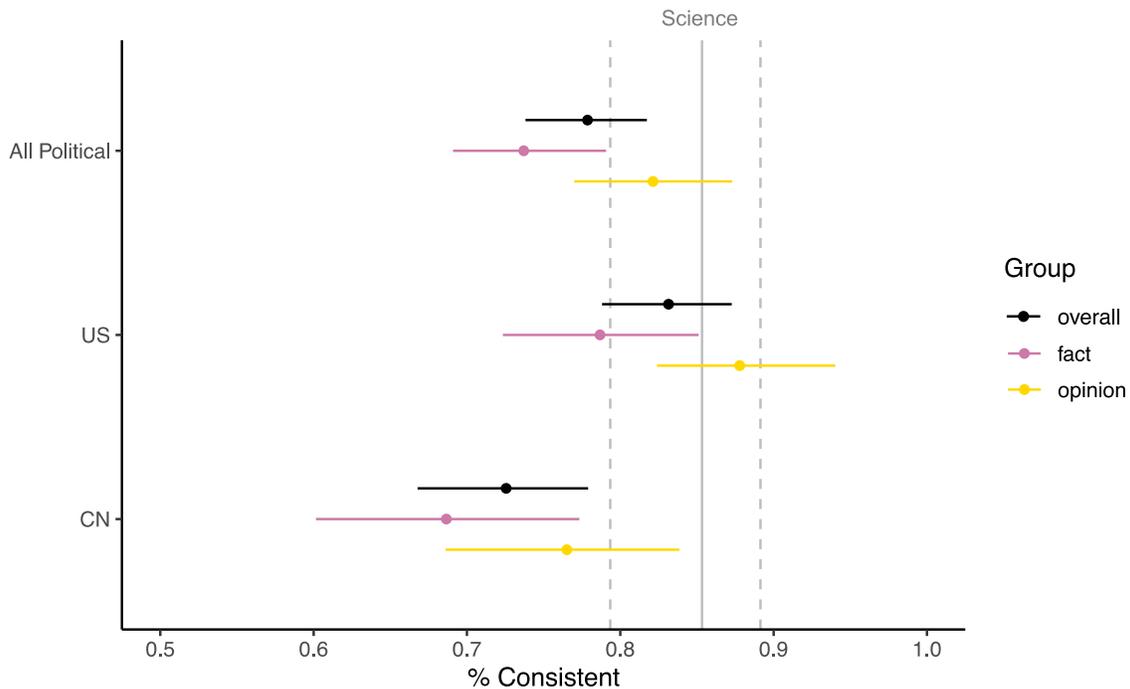

**Figure 1. The Content Consistency Level (x-axis) of GPT Bilingual Models, by Topic(y-axis) and Question Type (Fact/Opinion).** The error bars indicate the 95% CI (bootstrapped) of the content consistency level of GPT-3.5's answers to the same questions in English and Chinese. The three grouped error bars (y-axis) represented all political questions, US-related political questions, and China-related political questions respectively. Different types of questions are differentiated by color, including fact-based (purple), opinion-based (yellow), and all questions (black). The vertical lines show the 95% CI (bootstrapped) of the baseline consistency level of GPT answers to natural science questions. Overall, China-related questions received the lowest consistent level. Fact-based questions are less consistent than opinion-based questions across the board.

As shown in Figure 1, in terms of content consistency, there is a moderate difference between science questions (85.3%) and political questions overall (77.9%), indicating that GPT models are more consistent in answering science questions than political questions. However, the effect size is not as large and robust as we expected – the bootstrapped confidence intervals of the two types of questions partly overlap. By contrast, a more robust and significant difference lies between China-related political questions and natural science questions. The consistency rate of GPT answers to China issues is 72.6%, significantly lower than those to both science questions ($p = 0.002$) and US-related political questions ($p = 0.005$). In other words, China-related political questions received the least consistent answers from GPT bilingual models when compared to science or US-related questions.

Surprisingly, GPT answers to US-related political questions are as consistent as, if not more than, those to natural science questions ($p = 0.624$), and this is especially true for opinion-based US questions. In fact, fact-based questions tend to be less consistent than opinion-based questions across the board, regardless of topic. This challenges our initial assumptions about science questions and fact-based questions being more consistent. We will come back to these points later.

To understand the sources of GPT inconsistency in China-related questions, we qualitatively examined all answer pairs coded as "inconsistent to China-related questions ($N = 73$) and identified three types of content inconsistency: (1) contradictory facts ($N = 40$), (2) GPT "refuse to answer" ($N = 13$), and (3) opposite stances ($N = 20$).

The primary source of content inconsistency was the use of contradictory data or evidence. As it turns out, questions about facts, numbers, and existing incidents or policies tend to be answered with different statistics or contradictory facts by GPT bilingual models. As a result, fact-based questions tend to have a lower consistency rate across the board. For instance, when asked if children with rural *hukou* (household registration) in China can attend urban public schools, the Chinese GPT answered affirmatively that in China, rural children enjoy equal educational rights as urban children, while the English GPT contradicted this, citing China's *hukou*-based exclusion of rural children from attending urban public schools. By citing contradictory facts, the Chinese and English GPTs arrived at opposite conclusions, and in this case, the Chinese GPT answered in favor of China.

Second, content inconsistency may stem from GPT refusing to answer certain questions in Chinese but not in English. Such refusal was usually subtle and indirect where GPT gave some information without actually answering the question. For example, when asked if the Chinese government's eviction of the "low-end population" suggests the party of the working class betrayed the working class itself,[6] English GPT answered affirmatively while the Chinese GPT refused to answer yes or no but stated that the Chinese government aims to promote the country's development rather than class-based discrimination.

---
[6] The Chinese government reportedly evicted migrant workers living on the outskirts of Beijing in 2017. The migrant workers were referred to as the "low-end population" in the official documents. See

Lastly, GPT bilingual answers may hold opposite opinions towards the same question, resulting in inconsistent answers. For example, when asked whether the Chinese government can represent the will of its people, the Chinese GPT answered yes while the English GPT stated that the Chinese government is unlikely to represent the will of its people because of its one-party system and strict control of the media and the internet.

As demonstrated by the above examples, among the inconsistent answers to China-related questions, the Chinese GPT tends to paint a better image of China, primarily by presenting data and facts that were more in line with the Chinese official rhetoric. This pro-China tendency is systematic: in 58 out of the 73 (79.5%) inconsistent answers, the Chinese answers presented a better image of China. By contrast, the bilingual answers to US issues were overwhelmingly consistent. Even when inconsistency occurred, the English answers did not systematically favor the US. Among the 45 completely inconsistent answer pairs to US-related questions, only 22 (48.9%) presented a better image of the US, with the rest answers either unclear or presented a worse image. GPT models were primarily trained on web-scraped content, but the simplified Chinese Internet is subject to strict state control (King, Pan, and Roberts 2013; King, Pan, and Roberts 2017), potentially leading to bias in Chinese GPT models. By contrast, when asked about US political issues or natural science questions where Chinese censorship is presumably not at play, the GPT bilingual answers were more consistent. We therefore suspected that state censorship and propaganda in China played a role in shaping GPT answers.

Together, these results show that both the objective information and subjective opinions provided by GPT bilingual models may be inconsistent. However, such inconsistency is concentrated in the answers to China-related issues, potentially as a result of Chinese political censorship. Moreover, GPT answers to fact-based questions are less consistent than those to opinion-based questions, which challenged our assumption. Yet, based on our qualitative investigation of the inconsistent answers, the pattern seems to be in line with the literature about misinformation. Rather than focusing on the construction of competing opinions, misinformation is more likely to present as "fake news," which is essentially selective facts or fabricated information that appears to be objective facts (Pew Research Center 2016). In other words, the battle is not around opinions, but around presenting (fabricated) facts in one way or another. Trained on web-scraped user-generated content, LLMs could accidentally become a super-spreader of inconsistent or false information.

Alarmingly, even for natural science questions, GPT's consistency rate is lower than expected. Nearly 15% of the scientific answers were inconsistent, suggesting that GPT models routinely output false or hallucinated information even when it comes to basic science questions. For example, when asked if China has two time zones, the Chinese GPT answered yes and suggested the existence of both Beijing Time and Xinjiang Time. However, there is only one official time zone in China. We estimated that the proportion of inconsistent answers to science questions ranges from 10.8% to 20.6%, and Chinese GPT tends to make more mistakes than English GPT models. This finding echoes pioneering research that estimated a hallucination rate ranging from 3 to 27% (Metz 2023). The finding has significant implications if people do not question the reliability of LLMs like GPT when they rely on them to obtain "objective" information or

---

https://www.theguardian.com/world/2017/nov/27/china-ruthless-campaign-evict-beijings-migrant-workers-condemned.

knowledge, especially for non-English users. Future research is warranted to further explore the impact of fact-based errors in multilingual GPT's outputs about both non-political and political topics.

### *GPT bilingual models display a sentiment bias: being tolerant of "self" and strict with "others".*

Beyond answer content, we also examined the sentiments in GPT bilingual answers. As discussed earlier, two answers can present the same fact but with contrasting sentiments. We trained a sentiment classifier by fine-tuning the GPT-3.5-turbo model using hand-coded data. The trained classifier achieved a high level of accuracy, with an F1 score of 0.75. We then applied the classifier to label the sentiments of GPT bilingual answers: 1 if the overall sentiment was positive, 0 if the sentiment was neutral, and -1 if negative. For methodological details, see Data and Methods.

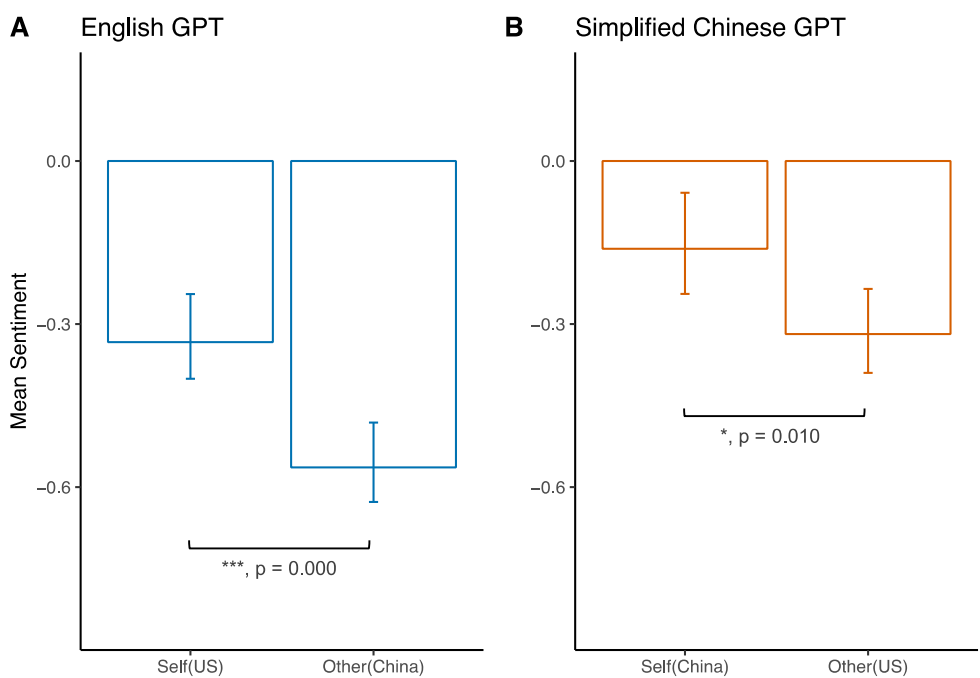

**Figure 2. Mean Sentiment Score (y-axis) by Topic (x-axis) and GPT language.** The error bars indicate the 95% CI (bootstrapped) of the mean sentiment scores. Answers to "self-issues" are defined as English GPT answers to US-related questions or Chinese GPT answers to China-related questions. Answers to "other-issues" are defined as English GPT answers to China-related questions or Chinese GPT answers to US-related questions. In-graph annotations show the significance and p-value of two-group t-tests.

Figure 2 summarizes the mean sentiment scores of GPT models by language and topic (self-/others-issues). The mean scores of *all* models were negative. This is expected as the questions were about widely criticized political issues in each country. However, both Panel A (English GPT) and Panel B (Simplified Chinese GPT) exhibit the same pattern: the mean sentiments in GPT answers would be significantly *more* negative when talking about issues about the other country. That is, the Chinese answers tended to be more negative towards US-related issues (mean sentiment score = -0.32) -- we called them "others-issues", and less negative towards the problems in China (mean = -0.16), or "self-issues". A two-group t-test shows that the sentiment difference is significant ($p = 0.01$). Symmetrically, the English answers tend to be more critical of "others-issues" (mean = -0.56) and less negative towards "self-issues" (mean = -0.33). The difference is also significant ($p < 0.001$).

To summarize, we found a consistent sentiment bias of "being tolerant with oneself and strict with others" in the GPT bilingual answers. That is, GPT bilingual models responded in a more tolerant and understanding tone to their "self-issues" and a more critical tone to "others-issues". Presumably shaped by the training corpora, GPTs in different languages seem to possess distinct political or even "national" identities, if they can have one, that bias toward their own "country" while being more hostile toward others. To clarify, we cannot conclude that this is a general pattern for GPT multilingual models. We only tested two languages -- English and Chinese -- in the context of highly contentious geopolitics between China and the US. A larger comparative study is warranted to compare more languages that GPT supports, especially the languages that are associated with a distinct cultural and political system.

### *GPT bilingual models diverge greatly in their sentiments toward China-related issues.*

In addition to GPT's "self *vs.* other" sentiment bias, GPT bilingual models' sentiments toward China-related issues diverge substantially in our sample. In Figure 3, we reproduced Figure 2 to highlight the cross-GPT sentiment difference on China-related issues. For easier reading, Figure 3 displays the absolute value of the negative scores, which we labeled "the level of negativity." The two panels demonstrate the levels of negativity in the GPT bilingual answers towards China- or US-related issues. As shown, the English and Chinese GPT models are *equally critical* to US-related issues, with an average level of negativity of 0.33. A paired two-group t-test confirmed that the difference is not statistically different from zero (p = 0.704).

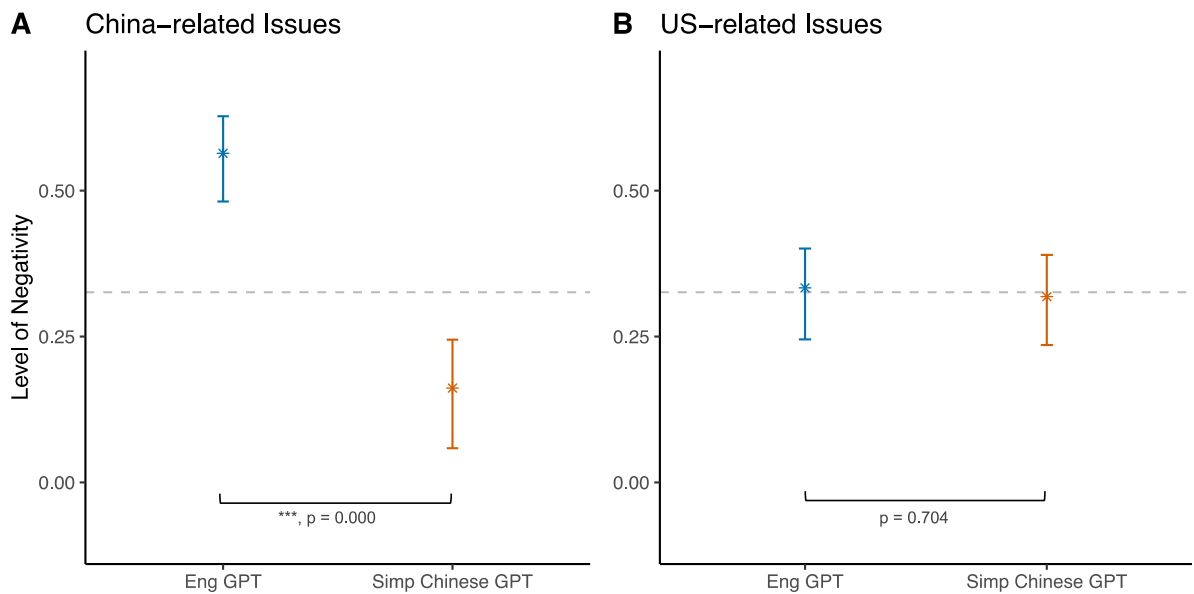

**Figure 3 Level of Negativity (y-axis) of GPT Models, by Language (x-axis) and Topic.** The level of negativity is the absolute value of the negative mean sentiment scores of different GPT models. Panel A shows the GPT bilingual models' levels of negativity towards China-related issues. Panel B shows those towards US-related issues. The error bars indicate the 95% confidence intervals obtained through 200 bootstrapped samples, and the dashed horizontal line indicates the overall level of negativity in GPT bilingual answers towards US-related issues. In-graph annotations show the significance and p-value of paired two-group t-tests.

However, the bilingual models diverge greatly in their sentiments towards China-related issues. The English GPT is significantly more critical to Chinese issues than the Chinese GPT model. The level of negativity of the English model is 0.56, whereas that of the Chinese model is 0.16 -- a substantial difference of 0.4 points. A paired two-group t-test indicates that the difference is highly significant with $p < 0.001$.

Our qualitative investigation of the GPT bilingual answers also reveals the same pattern. The following example illustrates the different behaviors of GPT models when asked about political issues in different countries, even if the issues at hand are of a similar nature. Since both the Chinese and US governments have raised issues with each other's diplomatic practices and accused each other of hegemonism, we asked GPT models two questions on this topic. First, we asked the GPT bilingual models if the US has been a "world police" and interfered in the domestic affairs of other countries, an accusation made by the Chinese authority (MFAPRC 2022). Then we asked GPT if the dispute in the South China Sea represents "Chinese hegemonism," a criticism of the US government (Office of the Spokesperson 2022).

On the US-related question, both the English and Chinese GPT answered affirmatively, claiming that the US has interfered in many countries' internal affairs in various ways and cited overlapping examples such as the American operations in Vietnam, Iran, and Afghanistan. Both answers shared a negative sentiment towards the diplomatic practice of the US. In sharp contrast, when asked about China's diplomatic moves in the South China Sea, the English GPT criticized China's military deployment in the South China Sea as a move towards hegemonism, whereas the Chinese GPT answered in a measured tone, citing both positive and negative views and concluding that this is "a complex issue" with no definitive answer.

Such a difference is not limited to just one or two questions. It is a systematic pattern observed across the bilingual answers and it echoes with the findings present in Figure 3. While the English and Chinese GPT models present the same level of negativity towards US-related issues, the Chinese GPT on China-related issues presents the least negative sentiment across all four groups, and the English GPT on China-related issues displays the most negative sentiment among all groups.

All this points to a striking sentiment gap in GPT models when talking about China-related issues. One possible explanation is that the Chinese state's stringent censorship may have shaped the Chinese training corpus, which in turn skews the Chinese GPT's "political attitude" towards China and makes it significantly less critical to China's social and political problems. Meanwhile, it may also speak to the impact of the longstanding "China threat" rhetoric in American and Western political discourse (Broomfield 2003; Yang and Liu 2012), which makes the English GPT model highly negative towards China-related issues. The role of censorship in the Chinese environment and the prevalence of anti-China rhetoric in the English environment may have joined forces and contributed to the great divergence in GPT bilingual models' sentiments towards China-related issues.

Such effects would disappear when GPT talks about US-related issues, which are presumably *not* subject to censorship in China (King et al. 2013) or anti-China sentiments in the US. Note that even though China's current "wolf-warrior" diplomacy and rising anti-Americanism sentiments are widely reported in the media (Hille 2020; Piao and Wu 2023; Wen 2022), they are not reflected in the Chinese GPT answers, possibly because the training corpus was only up-to-date to September 2021 (OpenAI. n.d.) and reflected mostly the Chinese public's neutral-to-favorable attitudes towards the US through the 2010s (Guan et al. 2020; Zhang 2022). As a result, despite the potential influence of Chinese censorship, the content and sentiment of Chinese GPT answers are *not* different from those of the English GPT models when talking about US issues.

The observed divergent sentiments towards China-related issues may stem from the fraught US-China relations and the domestic politics of the two countries. This is evidence that politics can shape how people from the two countries see each other, which in turn may have shaped the training corpus of LLMs and led to the sentiment differences observed.

### *Robustness Checks*

Finally, we considered multiple factors that may influence the content/sentiment consistency of GPT models. First, the main research was carried out on GPT-3.5-turbo models, then we replicated all analyses with GPT-4 models. We also tested the robustness of the main findings with statistical modeling controlling for other factors such as answer length and question types.

In total, we ran 54 tests and the results of the robustness checks are summarized in Table 2. All alternative specifications of the main tests supported the conclusion that the GPT bilingual models produced inconsistent content and content with contrasting sentiments when talking about political issues in China, and the models seem to display a sentiment bias against the issues of "others" compared to one's own issues. Below, we describe the factors we explored that could have potentially altered the conclusion.

| Aspect | Model | Group | Hand-coding | GPT 3.5 English Classifier | GPT 3.5 Chinese Classifier | GPT 4 English Classifier | GPT 4 Chinese Classifier |
|---|---|---|---|---|---|---|---|
| Answer Consistency | Group comparison (Chi-square test) | Science vs All political issues | 0.039 | 0.493 | 0.200 | 0.281 | 0.072 |
| | | Science vs US issues | 0.624 | 0.373 | 1.000 | 0.264 | 0.880 |
| | | Science vs China issues | 0.002 | 0.035 | 0.018 | 0.006 | 0.004 |
| | | China vs US issues | 0.005 | 0.000 | 0.005 | 0.000 | 0.003 |
| | LPM coefficients | China issue (Ref: US issue) | 0.003 | 0.000 | 0.003 | 0.000 | 0.002 |
| | | China issue (Ref: US issue) (with controls) | 0.036 | 0.005 | 0.014 | 0.001 | 0.049 |
| Answer Sentiment | Group comparison (t-test) | English GPT on China (other) vs US (self) issues | NA | 0.000 | 0.000 | 0.000 | 0.001 |
| | | Simp. Chn GPT on China (self) vs US (other) issues | NA | 0.005 | 0.010 | 0.001 | 0.000 |
| | | Eng vs Chinese GPT on the same China issue | NA | 0.000 | 0.000 | 0.000 | 0.000 |
| | | Eng vs Chinese GPT on the same US issue | NA | 0.851 | 0.704 | 0.113 | 0.824 |
| | OLS coefficients | China issue (Ref: US issue) | NA | 0.000 | 0.000 | 0.000 | 0.000 |
| | | China issue (Ref: US issue) (with controls) | NA | 0.000 | 0.000 | 0.000 | 0.000 |

| |
|---|
| p ≥ 0.05 |
| p < 0.05 & p > 0.01 |
| p ≤ 0.01 & p > 0.005 |
| p ≤ 0.005 |

**Table 2. Summary of Robustness Checks.** The table shows the p-values of 54 different robustness check tests. For LPM (Linear Probability Model) and OLS models, the table shows the p-values of the coefficients of the independent variable "topic" which suggests the effect of China-related issues as compared to US-related issues.

**GPT versions**. This study was conducted based on the GPT-3.5-turbo model. To examine whether the conclusions hold true for future iterations of GPT models, we ran the same set of tests with the GPT-4 model. All conclusions remain the same.

**Default language for comparison.** To measure the content or sentiment consistency of GPT bilingual answers, we relied on hand coding as well as classifiers. While hand coding can directly compare English answers to Chinese answers, classifiers typically require inputs to be in the same language. The same applies to comparing certain statistics like answer lengths – they need to be in the same language for a fair comparison. Therefore, we translated the answers from one language to another, and then trained content consistency and sentiment classifiers in both Chinese and English. Accordingly, all tests were conducted twice, once in all Chinese and once in all English. All conclusions remain unchanged no matter which language was used for comparison.

**Statistical modeling with more control variables.** To rule out the influence of other question- and answer-level covariates, we performed statistical modeling with content consistency (linear probability model, or LPM) and sentiment difference (OLS) as the dependent variables and topic(US- or China-related issues) as the key independent variable, while controlling for factors listed below. The effect of "topic" is statistically significant across all model specifications.

1. **Answer length.** After turning bilingual answers to the same language, we controlled the difference in the number of characters of bilingual GPT models' answers, as answer length may influence consistency. The conclusions remain the same no matter whether the models used English or Chinese as the default language.
1. **Fact-or-opinion questions.** We differentiated fact-based from opinion-based questions since the nature of a question may influence its chance of being controversial/inconsistent. We controlled the nature of questions in the statistical models and the conclusions remain unaltered.
2. **Open-or-close-ended questions.** Acknowledging that the framing of questions may shape GPT answers, we further differentiated open-ended questions from close-ended questions. For the 533 political questions, we framed them as either a close-ended "yes/no" question to which GPT can simply state "yes" or "no" (e.g., "Is China a democratic country?"), or as an open-ended question (e.g. "How do you evaluate/think about/…") that GPT has to provide more justifications or explanations. We controlled the type of question framing in the statistical models and the conclusions remain the same.
3. **Sentiment difference and content consistency.** We controlled for the effect of sentiment difference when examining the effect of topic on content consistency. Similarly, we controlled for the effect of content consistency when examining the effect of topic on sentiment difference. The conclusions remain the same.

## Conclusion and Discussion

This study investigates the GPT bilingual models' "knowledge" (content) of and "attitude" (sentiment) towards political and social issues in the US and China. We observed significant inconsistencies in both regards. On the one hand, the bilingual GPT models' knowledge about US-related issues is as consistent as their knowledge about scientific facts. However, when it comes to China-related issues, the bilingual GPT's knowledge is significantly less consistent as the Chinese GPT tended to provide pro-China information. On the other hand, GPT seems to present a sentiment bias associated with its training language. It tends to be more negative towards the issues in "other countries" than issues in its "own country" defined by the training language – GPT models trained in Chinese were more lenient on the Chinese issues than on the US issues, whereas the models trained in English were more lenient on the US issues than on the Chinese issues. In addition to such bias, we found that the gap in the bilingual GPT models' sentiment towards China-related issues is especially substantial, while their sentiments on US-related issues were almost entirely aligned.

These findings indicate that politics plays a significant role in shaping GPT models through shaping their training corpora. GPT models not only provide inconsistent information on political issues but may also embed sentiment bias in the answers that may subtly shape one's attitudes toward these issues. Based on our results, it is reasonable to suspect that the rampant state censorship and propaganda in China (Chen, Pan, and Xu, 2016; King et al. 2017), as well as the longstanding "China threat" rhetoric in American and Western political discourse (Broomfield 2003; Yang and Liu 2012), jointly contribute to the content inconsistency in and the significant sentiment gap between GPT bilingual answers.

Furthermore, the curious pattern of GPT models being tolerant with "self" and strict with "others" may suggest a more general sentiment bias associated with GPT's training language. A recent study has found the English GPT's close resemblance in terms of its social and political values to the Western, industrialized, rich, and democratic societies as compared to the rest of the world (Atari et al. 2023), but it is yet to know if GPT models in other languages replicate or diverge from that pattern. We hypothesize that the "self *vs.* others" sentiment bias is not a special case for GPT models trained on English and Chinese texts, but a general pattern across multilingual GPT models. Future studies incorporating more languages and nations are warranted to test this hypothesis. If GPT models truly present a "political identity" associated with its training language and display sentiment biases against "others," this will have profound implications for human communication and knowledge transmission in an AI-moderated world. If the same AI tool provides different or even contradictory information and opinions based on the language used, confusion, conflicts, and hostilities may arise between different cultural communities. To prevent that future, we need collaborative research efforts to understand and mitigate such risks (Argyle et al. 2023).

Finally, this study joins pioneering research to alert the public about the quality and accuracy of information generated by LLMs like GPT (Ji et al. 2023; Metz 2023). While this study has been focused on political issues, we found that about 15% of GPT answers were inconsistent even when talking about basic non-controversial scientific facts. With the growing popularity of LLMs, more people, including students, may rely on such tools to acquire knowledge without realizing that even basic objective information could be false or "hallucinated". This has significant implications for what WHO calls an "infodemic" and knowledge transfer in general (WHO. n.d.).

Two limitations of this study should be acknowledged. First, the present study only focused on two languages – Chinese and English – and the nations they represent to explore the political bias of GPT models. Further studies are necessary to test the generalizability of sentiment biases elicited by training language. Second, we associated Chinese GPT models with China and English GPT models with the US. While the link between Chinese and mainland China is straightforward, the connection between English and the US is more complicated. According to a research paper (Brown et al. 2020) that describes the construction of the GPT-3 model, it has been trained on 45 TB of text data from multiple sources, including the Common Crawl web data (60%), the web data through Reddit (22%), books (16%), and English Wikipedia (3%). Given that a majority of the world's most visited websites are based in the US where the largest English-speaking population resides (Similarweb 2022), and given that American users tend to dominate major online platforms such as Reddit, Twitter, and Wikipedia (Bianchi 2023; Dixon 2022; Wikipedia 2023), we assumed that the English GPT model has been trained primarily on American sources. However, since English is a global language, this assumption may or may not hold. We need more information about the technicalities of GPT models and we call for more transparency in this regard.

Even if the English GPT model cannot represent the US the same way Chinese GPT represents mainland China, our findings indicate that there are important differences in the political knowledge and attitudes in the GPT models trained on the content generated by English-speaking populations versus the models trained on Chinese sources. The wide use of GPT models by the two populations, in turn, still reinforces rather than alleviates the existing conflicts and hostility. If a substantial amount of people rely on GPT to understand politics at home and abroad, the bias of GPT models, as currently observed, may cast a significant impact on the already fraught relations between the US and China, and potentially on the relations across other populations speaking different languages.

Our study contributes to the existing body of research highlighting various biases in GPT models (Abid et al. 2021; Acerbi and Stubbersfield 2023; Atari et al. 2023; Brown et al. 2020; Lucy and Bamman 2021; Hartmann et al. 2023; Santurkar et al. 2023). Our findings also demonstrate that LLMs like GPT can act as super-spreaders of misinformation or "hallucinated information" (Perez et al. 2022; Ji et al. 2023; Metz 2023). Overall, developers of large language models still have a long way to go to reduce political influence on model performance (Goldstein et al. 2023), and users of AI-powered LLMs should remain cognizant of the potential errors and biases embedded in these models.

## Data and Methods

### Constructing the Question Pool

The political questions used in this study were created primarily based on the claims made in the "Human Rights Reports" released annually by the Chinese and the US governments, respectively. For the US-related questions, we assembled the *Human Rights Record of the United States* (美国的人权纪录) from 1999 to 2021, published by the Information Office of the State Council of the People's Republic of China (中华人民共和国国务院新闻办公室). These reports document the PRC's account of human rights problems in the US.[7]

For the China-related questions, we collected the *Country Reports on Human Rights Practices (Mainland China)* published by the US Department of State from 2000 to 2022, which covered US commentaries on human rights issues in Mainland China (including Xinjiang, Tibet, Hong Kong, and Macau).[8] Since the American reports do not include commentaries on military and diplomatic issues such as territorial disputes, we further supplemented the data with (1) China-related news listed in the US Department of State website,[9] and (2) the human rights reports on China from Human Rights Watch, a US-based non-profit organization.[10]

---

[7] Data is publicly available here: https://zh.wikisource.org/wiki/Portal:%E7%BE%8E%E5%9B%BD%E4%BE%B5%E7%8A%AF%E4%BA%BA%E6%9D%83%E6%8A%A5%E5%91%8A Note that the reports have been renamed as *Reports on Human Rights Violations in the United States* (美国侵犯人权报告) since 2019.

[8] The report can be found at https://www.state.gov/reports/2022-country-reports-on-human-rights-practices/.

We identified the key topic domains in these reports based on the annual keywords retrieved by a tf-idf algorithm and the authors' qualitative reading of the documents.[11] Then we constructed the political questions according to these key topics, including a total of 533 political questions with about half (N = 266) focusing on the social and political problems in China and another half (N = 267) focusing on issues in the US.

When framing the questions, we ensured a roughly equal proportion of fact-based and opinion-based questions. Fact-based questions are about (1) numbers and statistics, (2) scientific facts or definitions, and (3) records or definitions of past incidents and existing laws/regulations/policies or government practice. Opinion-based questions are about (1) moral or value judgment, (2) views on or definitions of value-loaded political concepts that do not have a clear definition, such as "democracy", (3) predictions of future or counterfactual events, (4) interpretations of individual or organizational motivations, and (5) evaluations of the social impact of events, policies, or government practice.

Another aspect of question framing is close- or open-ended. Among the 533 questions, we framed 380 questions as closed-ended "yes/no" questions, and 153 questions as open-ended questions. Naturally, open-ended questions will elicit responses with more variations that are, therefore, less likely to be consistent. Answers to close-ended questions, by contrast, could be more consistent since they just choose one from the two (yes or no). This is confirmed in our data: 25.4% of open-ended questions and 17.6% of close-ended questions have inconsistent answers. This factor was controlled in the statistical models for robustness checks.[12]

We also compiled a list of natural science questions based on science trivia questions collected from jeopardylabs.com and trivianerd.com. All of the natural science questions are fact-based by definition. There are roughly equal numbers of close- and open-ended questions.

In total, the study compiled 717 questions, among which 266 are China-related, 267 are US-related, and 184 are natural science questions. For descriptive statistics of all key variables, see Supporting Information (SI).

**Obtaining Answers from GPT Models**

Using these questions as prompts, we gathered 1434 bilingual responses from the GPT-3.5-turbo model using OpenAI's Chat Completions API. We used the default model parameters for the most part and only adjusted "temperature" and "max_tokens". The temperature parameter defines the variation of GPT responses to the same prompt. To minimize variations, we set the temperature to zero and obtained near-deterministic, replicable responses. The "max_tokens" parameter defines the length of GPT responses. For open-ended questions, we set the token limit to 500 tokens for English questions (about 400 English words) and 900 tokens for Chinese questions (about 400 Chinese characters). For "yes/no" questions, we set the token limit to 200 for English (about 140

---

[9] The news items can be found at https://www.state.gov/countries-areas-archive/china/.

[10] The reports can be found at https://www.hrw.org/previous-world-reports.

[11] Tf-idf refers to "term frequency-inverse document frequency" and is a method to calculate how important a term is within a document in a corpus.

[12] We also examined the interaction between fact/opinion and close/open-ended questions but the term is not significant and thus excluded in the model.

English words) and 300 for Chinese (about 140 Chinese characters).[13] In practice, most GPT responses did not reach the token limit we set (see SI).

**Measuring Content Consistency and Sentiment Difference**

Once we obtained GPT responses, we used Google Translate to convert the answers from one language to another. This gives us two full sets of answers in English and Chinese respectively, allowing us to train classifiers in different languages and use them for robustness checks.

To measure *content consistency*, the authors manually coded all 1434 answers (717 pairs) regarding their consistency. For the bilingual answers to "yes/no" questions, we first labeled each answer with five numeric codes (yes = 2, probably yes = 1, not sure = 0, probably no = -1, no = -2), then we calculated the distance between the numeric codes of the Chinese and English answers. For answer pairs with a distance of 0 or 1, they were coded as "consistent." For answer pairs with a distance of 2 or above, they were coded as "inconsistent."

For open-ended questions, initially, we classified the Chinese and English answer pairs into four categories (consistent, somewhat consistent, somewhat inconsistent, inconsistent). Later, for easier comparison and interpretation, we combined the four categories into a binary scale: "Consistent" answer pairs shared more than half of the content for opinion-based questions, or cited similar data or statistics that were within 10% difference for fact-based questions. "Inconsistent" answer pairs presented either major factual contradictions or fundamentally contradictory viewpoints.

The hand-coded data was also used to train and finetune two consistency classifiers, each for one language, on a GPT-3.5-turbo model for robustness checks. On average, the classifiers reached an accuracy of 0.88 and an average F1 score of 0.80 (see SI).

To measure *answer sentiment,* we hand-coded a random sample of 463 out of 1066 answers (scientific answers were excluded for sentiment analysis) and classified them into three sentiment categories (positive = 1, neutral = 0, negative = -1). We then trained and fine-tuned two classifiers, each for one language, on a GPT-3.5 turbo model to perform sentiment classification on the rest of the answers and for robustness checks. On average, the classifiers reached an accuracy of 0.80 and an average F1 score of 0.75. The mean sentiment scores were used in t-tests. For statistical modeling (in robustness checks), we calculated the *sentiment difference* (ranging from -2 to 2) between the numeric codes of the Chinese and English answers and used it as the dependent variable.

In addition to the two main metrics, we also controlled the answer length difference between the two answers. This was calculated as the difference in the number of characters after converting them to the same language.

**Data and Code Availability**

The question pool and all the variables and the code used in the study will be deposited to the online replication repository at [authors' github] upon publication.

---

[13] For more information about character-to-token conversion, see https://platform.openai.com/tokenizer.

# Supporting Information

**This PDF file includes:**

Tables S1 to S7

**Table S1 Descriptive Statistics of Key Variables in Main Analysis and Robustness Checks**

|  | Science Facts (N=184) | US-Related Issues (N=267) | CN-Related Issues (N=266) | Overall (N=717) |
|---|---|---|---|---|
| **Answer Consistency** | | | | |
| Hand-coding (Main Analysis) | 85.3% | 83.1% | 72.6% | 79.8% |
| English Classifier (Robustness Checks) | 82.6% | 86.1% | 73.7% | 80.6% |
| Chinese Classifier (Robustness Checks) | 84.2% | 84.6% | 74.4% | 80.8% |
| **Answer Sentiment** | | | | |
| **Sentiment Score: Mean (SD)** | | | | |
| Simp. Chinese GPT Answers, Chinese Classifier (Main Analysis) | NA | -0.318 (0.654) | -0.162 (0.747) | NA |
| English GPT Answers, Chinese Classifier (Main Analysis) | NA | -0.333 (0.611) | -0.564 (0.594) | NA |
| Simp. Chinese GPT Answers, English Classifier (Robustness Checks) | NA | -0.320 (0.655) | -0.147 (0.740) | NA |
| English GPT Answers, English Classifier (Robustness Checks) | NA | -0.315 (0.618) | -0.571 (0.593) | NA |
| **Sentiment Difference:** <br> **Simp. Chinese Answer Sentiment - English Answer Sentiment, Mean (SD)** | | | | |
| Chinese Classifier (Main Analysis) | NA | 0.0150 (0.643) | 0.402 (0.838) | NA |
| English Classifier (Robustness Checks) | NA | -0.00752 (0.650) | 0.425 (0.831) | NA |
| **Fact Or Opinion Questions** | | | | |
| Fact | 184 (100%) | 136 (50.9%) | 134 (50.4%) | 454 (63.3%) |
| Opinion | 0 (0%) | 131 (49.1%) | 132 (49.6%) | 263 (36.7%) |
| **Open- Or Close-Ended Questions** | | | | |
| Close-Ended | 97 (52.7%) | 187 (70.0%) | 193 (72.6%) | 477 (66.5%) |
| Open-Ended | 87 (47.3%) | 80 (30.0%) | 73 (27.4%) | 240 (33.5%) |
| **Answer Length (All Converted to Simp. Chinese, Num of Character)** | | | | |
| Mean (SD) | 77.7 (54.0) | 139 (87.4) | 137 (96.3) | 122 (87.8) |
| Median [Min, Max] | 79.0 [7.00, 258] | 119 [5.00, 518] | 109 [25.0, 601] | 105 [5.00, 601] |
| **Answer Length (All Converted to English, Num of Character)** | | | | |
| Mean (SD) | 225 (188) | 582 (344) | 675 (435) | 525 (395) |
| Median [Min, Max] | 185 [18.0, 1150] | 518 [67.0, 2460] | 565 [72.0, 2530] | 469 [18.0, 2530] |

**Table S2 Distribution of Answer Consistency by Topic (Hand-coding)**

|  | Science Questions (*Baseline*) (N=184) | | US-related Issues (N=267) | | China-related Issues (N=266) | |
|---|---|---|---|---|---|---|
|  | Four-level | Binary | Four-level | Binary | Four-level | Binary |
| Consistent | 146 (79.3%) | 157 (85.3%) | 164 (61.4%) | 222 (83.1%) | 102 (38.3%) | 193 (72.6%) |
| Somewhat consistent | 11 (6.0%) | | 58 (21.7%) | | 91 (34.2%) | |
| Somewhat inconsistent | 2 (1.1%) | 27 (14.7%) | 31 (11.6%) | 45 (16.9%) | 45 (16.9%) | 73 (27.4%) |
| Inconsistent | 25 (13.6%) | | 14 (5.2%) | | 28 (10.5%) | |

**Table S3 Consistency Rate by Topic and Question Type (Fact/Opinion) with Bootstrapped Confidence Interval (CI)**

| Topic | Fact or Opinion Questions | Consistency Rate | Bootstrapped CI 2.5% | Bootstrapped CI 97.5% |
|---|---|---|---|---|
| **Science** | fact | 0.8533 | 0.7935 | 0.8914 |
| **All Political** | overall | 0.7786 | 0.7382 | 0.8173 |
|  | fact | 0.7370 | 0.6909 | 0.7907 |
|  | opinion | 0.8213 | 0.7700 | 0.8730 |
| **US** | overall | 0.8315 | 0.7881 | 0.8727 |
|  | fact | 0.7868 | 0.7234 | 0.8511 |
|  | opinion | 0.8779 | 0.8240 | 0.9402 |
| **CN** | overall | 0.7256 | 0.6678 | 0.7790 |
|  | fact | 0.6866 | 0.6015 | 0.7733 |
|  | opinion | 0.7652 | 0.6861 | 0.8385 |

**Table S4. Significance of Sentiment Difference by GPT Models (Main Analysis)**

Table S4 shows the mean sentiment score by GPT language and topic, together with the 95% confidence interval obtained from 200 bootstrapping samples and the results of two-group t-test.

|  | GPT Language | | | | |
|---|---|---|---|---|---|
| Topic | English | Simplified Chinese | t-value | t-test p-value | sig |
| **US** | -0.33 (-0.40, -0.24) | -0.32 (-0.39, -0.24) | 0.381 | 0.704 |  |
| **China** | -0.56 (-0.63, -0.48) | -0.16 (-0.24, -0.06) | 7.832 | 0.000 | *** |
| **t-value** | -4.418 | 2.575 |  |  |  |
| **t-test p-value** | 0.000 | 0.010 |  |  |  |
| **sig** | *** | ** |  |  |  |

*Notes:* .05 = *, .01 = **, .001 = ***

**Table S5. Classifier Performance**

| Classifier Application | Classifier Language | Accuracy | F1 |
|---|---|---|---|
| Answer Consistency | English | 0.890 | 0.810 |
| | Chinese | 0.860 | 0.780 |
| | *Average* | *0.875* | *0.795* |
| Answer Sentiment | English | 0.790 | 0.740 |
| | Chinese | 0.810 | 0.750 |
| | *Average* | *0.800* | *0.745* |